\documentclass[10pt,twocolumn,letterpaper]{article}
\usepackage{cvpr}
\usepackage{times}
\usepackage[normalem]{ulem}
\usepackage[numbers]{natbib}
\usepackage[symbol]{footmisc}
\usepackage{amsmath}
\usepackage{amssymb}
\usepackage{array,bm,multirow}
\usepackage{array}
\usepackage{booktabs}
\usepackage[labelfont=bf]{caption}
\usepackage{color}
\usepackage{graphicx}
\usepackage{enumitem}
\usepackage{mathtools}
\usepackage{microtype}
\usepackage{multirow}
\usepackage{subfigure}
\usepackage{tabularx}
\usepackage{url}
\usepackage{xcolor}
\usepackage[pagebackref=true,breaklinks=true,colorlinks,bookmarks=false]{hyperref}
\usepackage{cleveref} 
\usepackage{lipsum}

\setlist{noitemsep,leftmargin=*}

\graphicspath{{./}{../}}

\cvprfinalcopy
\def\cvprPaperID{400}
\ifcvprfinal\fi

\useunder{\uline}{\ul}{}

\newcommand{\omitme}[1]{}
\newcommand{\pos}[1]{\textcolor{red}{#1}}
\newcommand{\negative}[1]{\textcolor{blue}{#1}}

\newcolumntype{L}[1]{>{\raggedright\arraybackslash}p{#1}}
\newcolumntype{C}[1]{>{\centering\arraybackslash}p{#1}}
\newcolumntype{R}[1]{>{\raggedleft\arraybackslash}p{#1}}

\newcommand\blfootnote[1]{%
  \begingroup
  \renewcommand\thefootnote{}\footnote{#1}%
  \addtocounter{footnote}{-1}%
  \endgroup
}

\makeatletter
\renewcommand{\paragraph}{%
 \@startsection{paragraph}{4}%
 {\z@}{0.5em}{-1em}%
 {\normalfont\normalsize\bfseries}%
}
\makeatother

\IfFileExists{/Users/vedaldi/.bash_profile}{
\hbadness=\maxdimen
\vbadness=\maxdimen
\vfuzz=30pt
\hfuzz=30pt
}{}

\title{Transferring Dense Pose to Proximal Animal Classes\vspace*{-3pt}}
\author{Artsiom Sanakoyeu\footnotemark\\
Heidelberg University
\and Vasil Khalidov\\
Facebook AI Research
\and Maureen S. McCarthy\\
MPI for Evolutionary Anthropology
\and Andrea Vedaldi\\
Facebook AI Research
\and Natalia Neverova\\
Facebook AI Research
}

\begin{document}
\twocolumn[\maketitle\vspace{-2em}    \centering
    \includegraphics[width=0.96\linewidth]{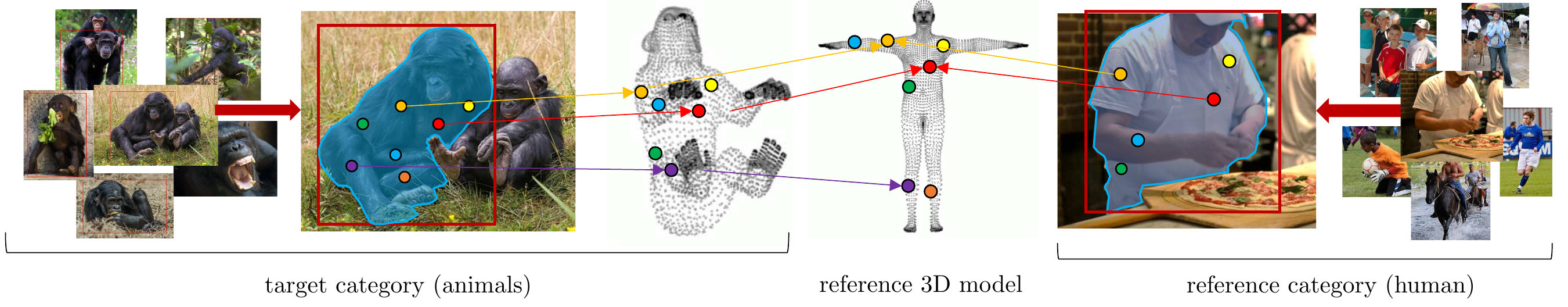}
    \vspace{-1em}
    \captionof{figure}{
    We consider the problem of dense pose labelling in animal classes. 
    We show that, for proximal to humans classes such as chimpanzees (left), we can obtain excellent performance by learning an integrated recognition architecture from existing data sources, including DensePose for humans as well as detection and segmentation information from other COCO classes (right).
    The key is to establish a common reference (middle), which we obtain via alignment of the reference models of the animals.
    This enables training a model for the target class without having to label a single example image for it.
    \label{fig:teaser}
}
\vspace{1em}

\medbreak]

\footnotetext[1]{Work done during an internship at Facebook AI Research}
\begin{abstract}
  Recent contributions have demonstrated that it is possible to recognize the pose of  humans densely and accurately 
given a large dataset of poses annotated in detail. 
  In principle, the same approach could be extended to any animal class, but the effort required for collecting new annotations for each case makes this strategy impractical, despite important applications in natural conservation, science and business.
  We show that, at least for proximal animal classes such as chimpanzees, it is possible to transfer the knowledge existing in dense pose recognition for humans, as well as in more general object detectors and segmenters, to the problem of dense pose recognition in other classes.
  We do this by (1) establishing a DensePose model for the new animal which is also geometrically aligned to humans (2) introducing a multi-head R-CNN architecture that facilitates transfer of multiple recognition tasks between classes, (3) finding which combination of known classes can be transferred most effectively to the new animal and (4) using self-calibrated uncertainty heads to generate pseudo-labels graded by quality for training a model for this class. 
  We also introduce two benchmark datasets labelled in the manner of DensePose for the class chimpanzee and use them to evaluate our approach, showing excellent transfer learning performance.%
\end{abstract}

\section{Introduction}\label{s:intro}

\blfootnote{Project page: \scalebox{1.0}{\href{https://asanakoy.github.io/densepose-evolution}{https://asanakoy.github.io/densepose-evolution}}}
In the past few years, computer vision has made significant progress in human pose recognition.
Deep networks can effectively detect and segment humans~\cite{he17mask}, localize their sparse 2D keypoints~\cite{newell2016stacked}, lift these 2D keypoints to 3D~\cite{novotny19c3dpo}, and even fit complex 3D models such as SMPL~\cite{kanazawa18end-to-end,kanazawa19learning}, all from a single picture or video.
DensePose~\cite{guler2018densepose} has shown that it is even possible to estimate a dense parameterization of pose by mapping individual image pixels to a canonical embedding space for the human body.

Such advances have been made possible by the introduction of large human pose datasets manually annotated with sparse or dense 2D keypoints, or even in 3D by means of capture systems such as domes.
For example, the DensePose-COCO dataset~\cite{guler2018densepose} contains 50K COCO images manually annotated with more than 5 millions human body points.
Clearly, collecting such data is very tedious, but is amply justified by the importance of human understanding in applications.
However, the natural world contains much more than just people.
For example, as of today scientists have identified 6,495 species of mammals, 60k vertebrates and 1.2M invertebrates~\cite{redlist2}.
The methods that have been developed for human understanding could likely be applied to most of these animals as well, provided that one is willing to incur the data annotation burden.
Unfortunately, while the applications of animal pose recognition in conservation, natural sciences, and business are numerous, just learning about one more animal may be difficult to justify economically, let alone learning about \emph{all} animals.

Yet, there is little reason to believe that these challenges are intrinsic.
Humans can understand the pose of most animals almost immediately, with good accuracy, and without requiring any data annotations at all.
Furthermore, images and videos of animals are abundant, so the bottleneck is the inability of machines to learn without external supervision.

In this paper, we thus consider the problem of learning to recognize the pose of animals with as little supervision as possible.
However, rather than starting from scratch, we want to make use of the rich annotations that are \emph{already} available for several animals, and humans in particular.
Thus, we focus on the problem of taking the existing annotated data as well as additional unlabelled images and videos of a target animal species and learn to recognize the pose of the latter.
Furthermore, for this study we restrict our attention to an animal species that is reasonably close to the available annotations, and elect to focus on the particular example of chimpanzees due to their evolutionary closeness to humans.\footnote{The idea is to eventually extend pose recognition to more and more animal species, in an incremental fashion.}
However, the findings in this paper are likely to generalize to many other classes as well.

We make several contributions in this work.
First, we introduce a dataset for chimpanzees, \emph{DensePose-Chimps}, labelled in the DensePose fashion, which we mostly use to assess quantitatively the performance of our methods.
We carefully design the canonical mapping for chimpanzees to be compatible with the one for humans in the original DensePose-COCO, in the sense that points in the two animal models are in as close a correspondence as possible. This is essential to be able to transfer dense pose recognition results from humans to chimpanzees while being able to asses the quality of the obtained results.

Second, we study in detail several strategies to transfer existing animal detectors, segmenters, and dense pose extractors from the available annotated data to chimpanzees.
In particular, while dense pose annotations exist only for humans, bounding box and mask annotations have been collected for several other object categories as well.
As a representative source dataset we thus consider COCO and we investigate how the different COCO classes can be combined to train an object detector and segmenter that transfers optimally to chimpanzees.
Surprisingly, we find that transfer from humans alone is not optimal, nor human is the best class for training a model for chimpanzees.
In addition to the DensePose-Chimps data, we collect human annotations for instance masks on the \emph{Chimp\&See}\footnote{Some of these videos are available at \url{http://www.zooniverse.org/projects/sassydumbledore/chimp-and-see}.} videos of chimpanzees captured with camera traps in the wild to evaluate the detection performance in the most challenging conditions (with severe occlusions, low visibility and motion blur).

Finally, we propose a framework for augmenting and adapting the human DensePose datasets to new species by self-supervision and pseudo-labeling with zero ground truth annotations on the target class.

\section{Related work}\label{s:realted}

\paragraph{Human pose recognition.}

There is abundant work on the recognition of human body pose, both in 2D and in 3D.
Given that our focus is 2D pose recognition, we discuss primarily the first class of methods.
2D human pose recognition has flourished by the introduction of deep neural networks~\cite{Wei2016,newell2016stacked,cao2017realtime} trained on large manually-annotated datasets of images and videos such as COCO~\cite{lin2014microsoft}, MPII~\cite{Andriluka-14}, Leeds Sports Pose Dataset (LSP)~\cite{Johnson10,Johnson11}, PennAction~\cite{zhang2013actemes} and Posetrack~\cite{PoseTrack}.
Furthermore, Dense Pose~\cite{guler2018densepose} has introduced a dataset with dense surface point annotations, mapping images to a $UV$ representation of a parametric 3D human model (SMPL)~\cite{Loper2015}.

While all such approaches are strongly-supervised, there are also methods that attempt to learn pose in a completely unsupervised manner~\cite{cliquecnn2016,thewlis17unsupervised,thewlis17bunsupervised,sanakoyeu2018pr,thewlis2019unsupervised,lorenz2019unsupervised,zhang2018unsupervised}.
Unfortunately, this technology is not sufficiently mature to compete with strong supervision in the wild.

\begin{figure*}[!t]
    \centering
    \includegraphics[width=\linewidth]{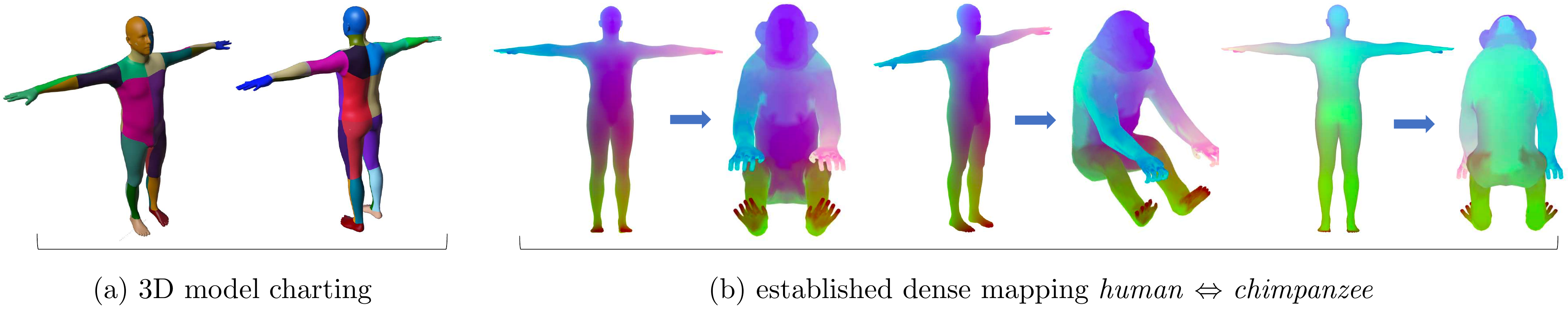}
    \caption{\label{fig:mapping} 3D shape re-mapping from the SMPL model for humans to new object categories (chimps). Manually defined semantic charting (a) on both models is used to establish dense correspondences (b) based on continuous semantic descriptors}
\end{figure*}

\paragraph{Animal pose recognition.}

Also related to our work, several authors have learned visual models of animals for the purpose of detection, segmentation, and pose recognition.
Some animals are included in almost all general-purpose 2D visual recognition datasets, and in COCO in particular.
Hence, all recent detectors and segmenters have been tested on at least a few animal classes.

For pose recognition, however, the existing body of research is more restricted.
Some recent papers have focused on designing pose estimation systems and benchmarks for particular animal species such as Amur tigers~\cite{li2019tigers}, cheetahs~\cite{nath2019cheetah} or drosophila melanogaster flies~\cite{Deepfly3d}.
There have been a number of large efforts on designing annotation tools for animals, such as DeepLabCut~\cite{deeplabcut} and Anipose~\cite{anipose}.
These tools also provide functionality for lifting 2D keypoints to 3D by using multiple views and triangulation.
A more detailed overview on applying computer vision and machine learning methodology in neuroscience and zoology is given in~\cite{mathis2019neuro}.
One of the main challenges in this field remains the narrow focus of existing research on specific kinds of animals and particular environments.

There have been few works focusing on the problem of animal understanding from visual data alone and in a more systematic way.
This includes the estimation of facial landmarks through domain adaptation~\cite{sheep2015,rashid2017interspecies}, and very recently full body pose estimation~\cite{cao2019animalpose} of four-legged animals by combining large-scale human datasets with a smaller number of animal annotations in a cross-domain adaptation framework.
Finally, a line of work from Zuffi et al.~\cite{zuffi2017menagerie,zuffi2018lions,zuffi2019safari} is exploring the problem of model-based 3D pose and shape estimation for animal classes.
Their research is based on parametric linear model, Skinned Multi-Animal Linear (SMAL), obtained from 3D scans of toy animals and having the capacity to represent multiple classes of mammals.
SMAL is the animal analogous of the popular SMLP~\cite{loper15smpl} model for humans.
It has since been used in other publications~\cite{biggs2018} for 3D animal reconstruction, but these methods may still be insufficiently robust for deployment in the wild.

\paragraph{Unsupervised and less supervised pose recognition.}

Recent methods such as {}\cite{thewlis17unsupervised,thewlis17bunsupervised,thewlis2019unsupervised,jakabunsupervised,zhang2018unsupervised,lorenz2019unsupervised} learn sparse and dense object landmarks for simple classes without making use of any annotation, but are too fragile to be used in our application.
Also relevant to our work, Slim DensePose~\cite{neverova19slim} looked at reducing the number of annotations required to learn a good DensePose model for humans.

\paragraph{Self-training for dense prediction.}

A recent study~\cite{yalniz2019billion} has demonstrated effectiveness of self-training on the task of image classification when scaled to large amounts of unlabeled data.
Pseudo-labeling by averaging predictions from multiple transformed versions of unlabeled samples has been shown effective for keypoint estimation~\cite{radosavovic2018omni}.
However, there has been very little research on self-training in the context of dense prediction tasks.
A recent work~\cite{SaltBodiesSegmentation2019} explored the idea of self-training for segmentation of seismic images and showed promising results on this task for the first time.


\section{Method}\label{s:method}

We wish to develop a methodology to learn Dense Pose models for new classes with minimal annotation effort.
Existing labelled datasets for object detection, segmentation and pose estimation, provide a significant source of supervision that can be harnessed for this task.
For detection and segmentation, COCO provide extensive annotations for a variety of object classes, including several animals.
For pose recognition, however, the available supervision is generally limited to humans, with a few exceptions.
Furthermore, for \emph{dense} pose recognition only human datasets are available --- the best example of which is DensePose-COCO~\cite{guler2018densepose}.

In this work, we raise a number of questions most critical for this setup, namely:
\begin{itemize}
\item defining learning and evaluation protocols on new animal categories allowing for training class-specific or class-agnostic DensePose models on a variety of species in a unified way (described in Sect.~\ref{s:combo});
\item improving quality of DensePose models and their robustness to unseen data distributions at test time (discussed in Sect.~\ref{s:multihead} and \ref{s:calibrated});
\item optimally combining the existing variety of data sources in order to initialize a detection model for a new animal species (discussed in Sect.~\ref{s:transfer});
\item defining strategies for mining dense pseudo-labels for gradual domain adaptation from humans to chimpanzees in a teacher-student setting (discussed in Sect.~\ref{s:distillation}).
\end{itemize}

\subsection{Annotation through 3D shape re-mapping}\label{s:combo}


While our aim is to learn to reconstruct the dense pose of chimpanzees with zero supervision, a manually-annotated dataset for this class is required for evaluation.
Here, we explain how to collect DensePose annotations for a new category, such as chimpanzees.


\paragraph{Dense Pose model.}

Recall that DensePose-COCO contains images of people collected `in the wild' and annotated with dense correspondences.
These dense keypoints are identified as the point $p\in S$ of a reference 3D model $S \subset \mathbb{R}^3$ of the object.\footnote{Dense Pose uses SMPL~\cite{Loper2015} to define $S$ due to its popularity}
Furthermore, the keypoints $p \in S$ are indexed by triplets $(c,u,v) \in \mathcal \{ 1,\dots,C\}\times[0,1]^2$ where $c$ is the \emph{chart index}, corresponding to one of $C$ model parts, and $(u,v)$ are the coordinates within a chart.
The DensePose-COCO dataset~\cite{guler2018densepose} contains bounding boxes, pixel-perfect foreground-background and part segmentations, and $(c,u,v)$ annotations for a large number of foreground pixels.

\paragraph{Dense Pose for chimps.}

We wish to extend the DensePose annotations to the chimpanzee class.
In order to do so, we rely on a separate artist-created 3D model\footnote{Purchased from http://hum3d.com/} of a chimpanzee as a reference for annotators to collect labels for the chimpanzee images (instead of the human model used by the original DensePose).

For each object, we use Amazon Mechanical Turk to collect the object bounding boxes, followed by pixel-perfect foreground/background segmentation masks, and finally the $(c,u,v)$ chart coordinates for a certain number of pixels randomly sampled from the foreground regions.
Differently from the original DensePose, we \emph{do not} also collect dense annotations for the body parts as the latter was found to be very challenging for the annotators.
Note however that the chart index $c$ reveals the part identity for each of the annotated image pixels.


\paragraph{Semantic alignment.}

Finally, we wish to align the human and chimpanzee DensePose models by mapping the collected annotations back on the surface of the SMPL model using the mesh re-mapping strategy described below.
The latter step unifies the evaluation protocols across different object categories and allows to transfer knowledge and annotations between different species.

In spite of the fact that humans and most mammals share topology and the skeletal structure, establishing precise semantic dense correspondences between the 3D models of humans and different animal species is challenging due to differences in body proportions and local geometry.

As preprocessing, we manually charted the SMPL and the chimp meshes into $L=32$ semantically-corresponding parts to guide the mapping.
Then, for each vertex $p$ of each mesh $S$, we extracted an adapted version of the continuous semantic descriptor $\mathbf{d}(p)$ proposed by L\'eon et al.~\cite{leon2016semantic}:
\begin{equation}
\mathbf{d}(p)= \left( d_{\ell}(p) \right)_{\ell=1}^L,
~~~
d_{\ell}(p)=\frac{1}{|S_{\ell}|}\sum _{s \in S_{\ell}}g(p, s;S_\ell)
\end{equation}
where $S_{\ell} \subset S$ is the set of all vertices in part $\ell$ of the mesh and $g(p,s)$ is the geodesic distance between two points on $S$.\footnote{%
To partially compensate for differences in proportions across different categories, we further normalized the descriptors by their part average:
$
  d_{\ell}(p) \leftarrow
  d_{\ell}(p) /
  \langle
  d_{\ell}(q)
  \rangle_{q\in S_\ell}.
$
}
With this, the mapping from the human mesh $S$ to the chimp mesh $S'$ is obtained by matching nearest descriptors:
$S \rightarrow S'$,
$p \mapsto \operatornamewithlimits{argmin}_{q\in S'} \|\mathbf{d}_S(p) - \mathbf{d}_{S'}(q)\|^2$.


This simple approach yields satisfactory results both in terms of alignment and smoothness, as shown in Fig.~\ref{fig:mapping}.
It does not require any optimization in 3D space based on model fitting or mesh deformation and works on meshes of arbitrary resolutions.
Interestingly. exploiting information about mesh geometry (such as high dimensional SHOT~\cite{salti2014shot} descriptors or their learned variants~\cite{halimi2019}) instead or in addition to semantic features results in noisy mappings.
This can likely be attributed to prominent inconsistencies in local geometry of some body regions between the object categories.


\begin{figure}[t!]
\begin{center}
   \includegraphics[width=1.0\linewidth]{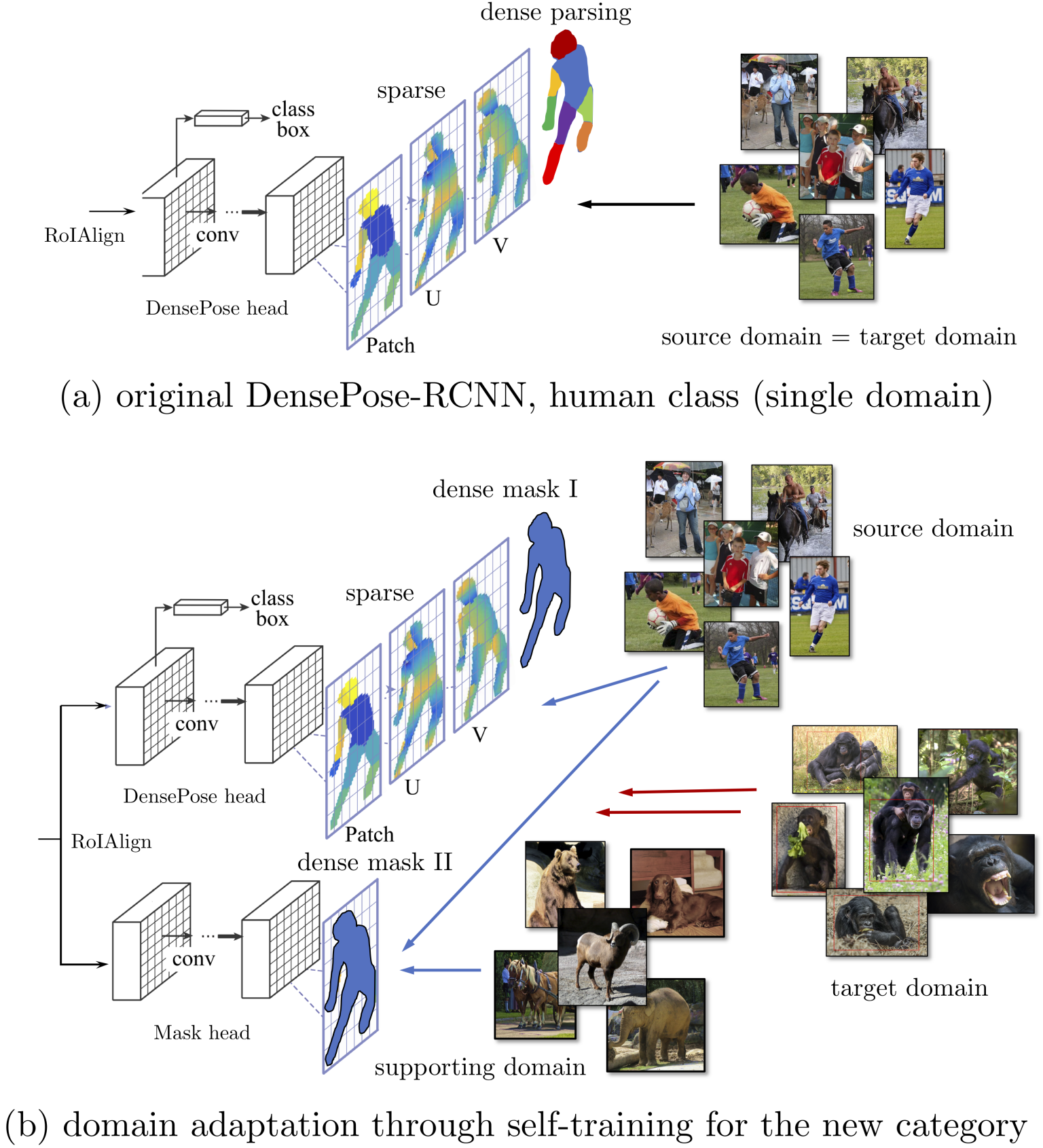}
\caption{\label{pipeline} Comparison of the original (a) and our (b) DensePose learning architecture.
See Sect.~\ref{s:multihead} for detailed description of the architecture.}
\end{center}
\end{figure}

\subsection{Multi-head R-CNN}\label{s:multihead}


Our goal is to develop a DensePose predictor for a new class.
Such a predictor must detect the object via a bounding box, segment it from the background, and obtain the DensePose chart and $uv$-map coordinates for each foreground pixel.
We implement this with a \emph{single model} with multiple heads, performing the various tasks on top of the same trunk and shared image features (Fig.\ref{pipeline}.b).

The base model is R-CNN~\cite{he17mask} modified to include the following heads.
The first head refines the coordinates of the bounding box.
The second head computes a foreground-background segmentation mask in the same way as Mask R-CNN\@.
The third and the final head computes a part segmentation mask $I$, assigning each pixel to one of the $24$ Dense Pose charts, and
the $uv$ map values for each foreground pixel.

\paragraph{Class-agnostic model.}

Compared to the standard Mask R-CNN, our model is \emph{class agnostic}, i.e.~trained for only one class type.
This is true also when we make use of a Mask R-CNN pre-trained on multiple source classes as the goal is always to only build a model for the final target chimpanzee class --- we found that merging classes is an effective way of integrating information.

\paragraph{Heterogeneous training.}

Our training data can be heterogeneous.
In particular, COCO provides segmentation masks for 80 categories, but DensePose-COCO provides DensePose annotations only for humans.
While we train a single class-agnostic model, the Dense Pose head is trained only for the class human for which the necessary ground-truth data is available.

Note in particular that both the Mask R-CNN head and the DensePose head contain a foreground-background segmentation component --- these are not equivalent, as the DensePose one is only valid (and trainable) for humans, while the Mask R-CNN one is generic (and trainable from all COCO classes).
We will see in the experiments that their combination improves performance.



\paragraph{Fine-tuning.}

As shown later, for fine-tuning the model we generate pseudo-label on chimpanzees imagery.
The pseudo-labels are generated for all components of the model (segmentations, $uv$ maps), including in particular both foreground-background segmentation heads.

\paragraph{Other architectural improvements.}

Our model (Fig.~\ref{pipeline}.a) has a few mode differences compared to the original Dense Pose (Fig.~\ref{pipeline}.b) which we found useful to improved accuracy and/or data collection efficiency.

First, both the original and our implementations use dense (pixel-perfect) supervision for the foreground-background masks.
However, in our version we \emph{do not} use the pixel-perfect part segmentations in the original DensePose annotations --- the part prediction head is trained only from the chart labels for the pixels that are annotated in the data.
This is another reason why we do not collect pixel-perfect segmentations for the chimpanzee images.


We further improve the DensePose head by implementing it using Panoptic Feature Pyramid Networks~\cite{kirillov2019panoptic}, and use a configuration similar to DeepLab~\cite{DeepLabV3} that benefits from higher resolution.


\subsection{Auto-calibrated R-CNN}\label{s:calibrated}

As suggested above, pseudo-labelling can be used to fine-tune a pre-trained model on imagery containing the target class, chimpanzees in our case.
The idea is to use a model pre-trained on a different class or set of classes to generate labels in the new domain, and then to retrain the model to fit those labels.
Due to the domain gap, however, the pseudo-labels are somewhat unreliable.
In this section, following~\cite{kendall17what} we develop a principled manner to let the neural network itself produce a \emph{calibrated measure of uncertainty} which we can use to rank pseudo-labels by reliability.




\paragraph{Classification uncertainty.}

Our model performs categorical classification for two purposes:
to associate a class label to a bounding box, and to classify individual pixels as background, foreground, or as one of the body parts.
In order to estimate the uncertainty for these categorical predictions, we adopt the \emph{temperature scaling} technique of~\cite{hinton2015distill}.

Thus let $z_{y}$ be the score that the neural network associates to hypothesis $y\in \{1,\dots,K\}$ for a given input sample.
We extend the network to compute an additional per-sample scalar $\alpha \geq 0$.
With this scalar, the posterior probability of hypothesis $y$ is given by the \emph{scaled softmax}
\begin{equation}
\hat{\sigma}
\left(y ; z, \alpha\right)
=
\frac
{\exp \left(\alpha z_{y}\right)}
{\sum_{k=1}^{K} \exp \left(\alpha z_{k}\right)}
\end{equation}
We can interpret the coefficient $\alpha=1/T$ as an inverse temperature.
A small $\alpha$ means that the model is fairly certain about the prediction, whereas a large $\alpha$ that it is not.

Note that, since $\alpha$ is also estimated by the neural network, we require a mechanism to learn it.
This is in fact obtained automatically~\cite{hinton2015distill,neumann2019relaxed} by simply minimizing the negative log-likelihood of the model, also known in this case as cross-entropy loss:
$
\ell(y, z, \alpha)
=
- \log \hat{\sigma} \left(y ; z, \alpha\right).
$

\begin{figure}[t!]
\centering
   \includegraphics[width=1.0\linewidth,height=0.3\linewidth]{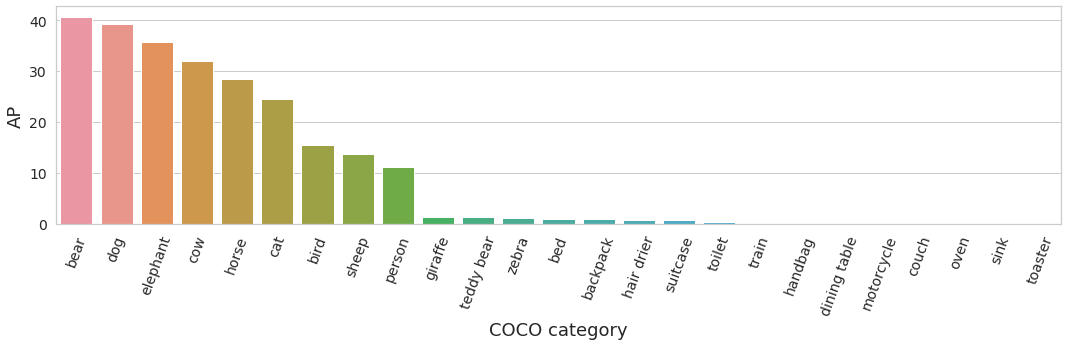}
   \vspace{-2em}
\caption{Instance Segmentation score (AP) on DensePose-Chimps for Mask R-CNN models trained using different COCO categories, ranked by decreasing performance.}\label{fig:coco_singe_category_ap}
\end{figure}

\paragraph{Regression with uncertainty.}

Our model performs regression to refine the bounding box proposals (for four scalar outputs, two for each of the two corners of the box) and to obtain the DensePose $uv$-coordinates (for two scalar outputs for each image pixel in a proposal).

Thus let $y\in\mathbb{R}^{D}$ be the vector emitted by one of the regression heads (where $D$ depends on the head).
Similarly to the classification case, we use the network to also predict an \emph{uncertainty score} $\sigma\in\mathbb{R}^D$.
This time, however, we have a different scalar for each element in $y$ (hence, for the $uv$-maps, we have two uncertainty scores for each pixel, which we can visualize as an image).
The vector $\sigma$ is interpreted as the diagonal variance of the regressed vector $y$, assuming the latter to have a Gaussian distribution.
The uncertainty scores $\sigma$ can thus be trained jointly with the predictor $\hat y$ by minimizing the negative log-likelihood of the model:
\begin{equation}\label{e:uregress}
\ell(y, \hat y, \sigma) \!=\!
\frac{D}{2}\!\log 2\pi +
\!\frac{1}{2}
\sum_{i=1}^D \!\left(
\log \sigma_i^2 +
\frac{(\hat y_i - y_i)^2}{\sigma^2_i}
\right)
\end{equation}
For a fixed error $|\hat y_i - y_i|$, the quantity above is minimized by setting $\sigma_i = |\hat y_i - y_i|$ --- hence the model is encouraged to guess the magnitude of its own prediction error.
However, if $|\hat y_i - y_i| =0$, the quantity above diverges to $-\infty$ for $\sigma_i\rightarrow 0$.
Hence, we clamp $\sigma_i$ from below to a minimum value $\sigma_\text{min} > 0$.

\begin{table*}[!htb]
\minipage{0.5\textwidth}
\centering
\scalebox{0.72}{
\begin{tabular}{l|ccc}
 model & $\textbf{AP}$ & $\textbf{AP}_{50}$ & $\textbf{AP}_{75}$  \\
 \midrule
DensePose-RCNN   & 50.88 &	80.40&	54.80 \\
DensePose-RCNN*  & 51.44&	81.44&	55.12 \\
DensePose-RCNN* ($\sigma$) & 54.13 &	82.32	& 58.06\\
\bottomrule
\end{tabular}}
\label{fig:SMPLFY}
\endminipage\hfill
\minipage{0.5\textwidth}
\centering
\scalebox{0.72}{
\begin{tabular}{l|ccc}
 model & $\textbf{AP}$ & $\textbf{AP}_{50}$ & $\textbf{AP}_{75}$  \\
 \midrule
 DensePose-RCNN   & 43.84 &	76.88 &	45.84 \\
 DensePose-RCNN*  & 43.84 &	77.52 &	45.60 \\
 DensePose-RCNN* ($\sigma$) & 45.58 &	78.79 &	47.93\\
\bottomrule
\end{tabular}
}
\endminipage
\caption{\label{tab:auto_calibrated_dp_rcnn} Detection (left) and instance segmentation (right) performance on DensePose-COCO \texttt{minival}.}

\end{table*}

\begin{table*}[!htb]
\centering
\scalebox{0.72}{
\begin{tabular}{l|ccc|cc|ccc|cc}
 model & $\textbf{AP}$ & $\textbf{AP}_{50}$ & $\textbf{AP}_{75}$ & $\textbf{AP}_{M}$ & $\textbf{AP}_{L}$ & $\textbf{AR}$ &
 $\textbf{AR}_{50}$ & $\textbf{AR}_{75}$ & $\textbf{AR}_{M}$ &
 $\textbf{AR}_{L}$\\
 \midrule
 DensePose-RCNN & 46.8&				84.5&	47.7&	41.8&	48.0 & 54.7 & 89.5 & 58.9 & 43.3 & 55.5\\
 DensePose-RCNN* & 47.2&				85.8&	47.3&	42.5 & 48.4 & 55.2 & 91.0 & 59.1 & 44.0 & 55.9	\\
 DensePose-RCNN* ($\sigma$) & 53.2&				88.3&	57.0 &	48.6&	54.6& 61.2 & 92.4 & 67.2 & 50.0 & 61.9\\
\bottomrule
\end{tabular}}
\caption{DensePose performance on DensePose-COCO \texttt{minival}. 
* denotes our improved architecture; ($\sigma$) denotes the proposed Auto-calibrated version of the network.}
\label{tab:auto_calibrated_dp_rcnn_2}
\end{table*}

\paragraph{Details.}

For both classification and regression models, the uncertainties $\alpha$ and $\sigma$ must be positive --- in the network, they are obtained via a \texttt{softplus} activation.

\subsection{Optimal transfer support}
\label{s:transfer}

In this section, we investigate which object categories in the COCO dataset provide the best support for recognizing a new animal species, chimpanzees in our case.
Among the animals in COCO, chimpanzees are most obviously related to humans, and we may thus expect that people may be the most transferable class.
However, despite their overall structural similarity, people's appearance is fairly different, also due to the lack of fur and the presence of clothing.
Furthermore, context is also often quite different.
It is thus unclear if a deep network trained to recognise humans can transfer well at all on chimpanzees, or whether other object categories might do better.\medskip

\noindent\textbf{Class selection.} We test what is more important: biological proximity of the species (as a proxy to morphological similarity) or appearance similarity (as a combination of typical poses and textures). We also search for a brute force solution for this particular dataset to back up or disprove our intuition for class selection. In our experiments, we have tested the following selections:
\begin{itemize}
    \item \emph{person} class only (due to morphological similarity).
    \item \emph{animal} classes only (due to higher pose and texture similarity): \emph{bear}, \emph{dog}, \emph{elephant}, \emph{cat}, \emph{horse}, \emph{cow}, \emph{bird}, \emph{sheep}, \emph{zebra}, \emph{giraffe}, \emph{mouse}.
    \item \emph{top-N} scoring classes on the new category (brute force solution). In this setting, we first train a set of $C$ single-class models for each of the $C=90$ object classes in the COCO dataset and rank them according to their instance segmentation performance on the DensePose-Chimps dataset (see Fig.~\ref{fig:coco_singe_category_ap}). Then for each combination of $S\in\{1,\ldots,C\}$ top scoring classes we train the same network from scratch. The solution that have we found optimal corresponds to $C_{\text{opt}}=9$, where the top-C scoring classes are: \emph{bear}, \emph{dog}, \emph{elephant}, \emph{cat}, \emph{horse}, \emph{cow}, \emph{bird}, \emph{person}, \emph{sheep}.

\end{itemize}
As shown in Tab.~\ref{t:transfer}, the top-N solution produces similar results compared to combination \emph{person+animals}. \emph{Person} class only is ineffective for training in this setting.

\noindent\textbf{Class fusion.}
We have also explored the question of class-agnostic vs multi-class training as a trade-off between the number of training samples per class vs granularity of prediction modes. For the task of adapting the new model to a single category (on the given dataset) class-agnostic training showed convincingly stronger results (see Tab.~\ref{t:transfer}).

\subsection{Dense label distillation}
\label{s:distillation}

Finally, we aim at finding an effective strategy for exploiting unlabeled data for the target domain in the teacher-student training setting and performing \emph{distillation} in dense prediction tasks.
In our setting, the \emph{teacher} network trained on the selected classes of the COCO dataset with DensePose is used to generate \emph{pseudo-labels} for fine-tuning the \emph{student} network 
on the augmented data. The \emph{student} network is initialized with \emph{teacher}'s weights.

Once teacher predictions on unlabeled data are obtained, we start by 
filtering out low confidence detections using calibrated detection scores. 
After that, the bounding boxes and segmentation masks on remaining samples are used for augmented training. For mining DensePose supervision, we consider three different dense sampling strategies driven by each of the tasks solved by the teacher network, in addition to uniform sampling:

\begin{itemize}
    \item \textbf{uniform sampling} -- all points from the selected detections are sampled with equal probability;
    \item \textbf{coarse classification uncertainty [mask-based]} -- sampling top $k$ from ranked calibrated posteriors produced by the mask branch for the task of binary classification;
    \item \textbf{fine classification uncertainty [$I$-based]} -- selection of top $k$ from ranked calibrated posteriors from the 24-way segmentation outputs of the DensePose head;
    \item \textbf{regression uncertainty sampling [$uv$-based]} -- sampling of top $k$ points based on ranked confidences in the $uv$-outputs of the DensePose head.
\end{itemize}

In Sect.~\ref{s:experiments} we provide experimental evidence that sampling based on confidence estimates from fine-grained tasks ($I$-estimation, $uv$-maps) results in the best \emph{student} performance.

\begin{table}[!t]
\centering
\scalebox{0.72}{
\begin{tabular}{l|c|ccc|cc} 
\multicolumn{2}{c}{\phantom{}} & \multicolumn{3}{c}{DensePose-Chimps} & \multicolumn{2}{c}{Chimp\&See} \\
\midrule
 sampling & $k$ &$\textbf{AP}_{DPose}$  & $\textbf{AP}_{D}$ & $\textbf{AP}_{S}$ &  $\textbf{AP}_{D}$ & $\textbf{AP}_{S}$ 
 \\
 \midrule
 --  & --  & 33.4 & 62.1 &	56.4 & 50.5 &	43.5 \\ 
 \midrule
 uniform    &5 & $34.5 \pm .4$ & $63.3 \pm .3$ & $58.0 \pm .3$ & $58.9 \pm .5$ & $49.0 \pm .5$ \\
 mask-based& 5 & $34.7 \pm .4$ & $63.3 \pm .3$ & $58.0 \pm .2$ & $58.8 \pm .6$ & $49.0 \pm .5$ \\
 $I$-based  & 5 & $\mathbf{34.9 \pm .6}$ &	$\mathbf{63.4 \pm .3}$ & $\mathbf{58.0 \pm .2}$ &	$\mathbf{59.2 \pm .4}$ &	$49.2 \pm .5$ \\
$uv$-based  & 5 & $34.6 \pm .3$ & $63.3 \pm .3$  &	$58.2 \pm .3$ &	$59.0 \pm .1$ &	$\mathbf{49.6 \pm .1}$ \\ 
\bottomrule
\end{tabular}}
\caption{\label{selftraining_strategy}AP of the \emph{student} network trained with different sampling strategies. Optimal number of sampled points $k$  per detection is reported for each sampling. The first row corresponds to the \emph{teacher} network. \textit{Mean}$\scriptstyle\pm$\textit{std} for $20$ runs.
}
\end{table}

\section{Experiments}\label{s:experiments}

We now describe the results of empirical evaluation and provide detailed descriptions of ablation studies.

\subsection{Datasets}

We use a combination of human and animal datasets with different kinds of annotations or no annotations at all. A brief description of each of them is provided below.

\paragraph{DensePose-COCO dataset~\cite{guler2018densepose}.} This is the dataset for human dense pose estimation, that we use for training the teacher model. It contains 50k annotated instances totalling to more than 5 million ground truth correspondences. We also augment the teacher training with other object categories from the original \textbf{COCO dataset}~\cite{lin2014microsoft}.

\paragraph{Chimp\&See dataset.} 
For training our models in a self-supervised setting, we used unlabeled videos containing chimpanzees from the \emph{Chimp\&See} project\footnote{A subset of the videos from the Chimp\&See dataset is publicly available at \url{http://www.zooniverse.org/projects/sassydumbledore/chimp-and-see}.}.
This data is being collected under the umbrella of The Pan African Programme\footnote{\url{http://panafrican.eva.mpg.de}}: The Cultured Chimpanzee (PanAf) by installing camera traps in more than 40 natural habitats of chimpanzees on different sites in Africa.
In this work, we used a subset of the collected data consisting of 18,\!556 video clips, from 10 sec to 1 min long each, captured with cameras in either standard or night vision mode depending on lighting conditions. 
These recordings were motion triggered automatically by passing animals. As a result, some clips may not contain any chimps beyond first several frames. 

For evaluation, we chose videos from one site, sampled frames at 1 fps, removed the near duplicates and collected human annotations for instance masks.
This resulted into 1054 images containing 1528 annotated instances, that we use to benchmark detection performance in our models.
However, due to in-the-wild nature of this data and presence of motion blur, severe occlusions, and low resolution in some cases, we found it infeasible to collect precise human annotations at the level of dense correspondences.

\paragraph{DensePose-Chimps test set.}
For the task of evaluating DensePose performance on this new category, we collected a set of $662$ higher quality images from Flickr that contain $933$ instances of chimpanzees. We annotated this data with bounding boxes, binary masks, body part segmentation and dense pose correspondences as explained in Sect.~\ref{s:combo}.

\begin{table}[!t]
\centering
\scalebox{0.72}{
\begin{tabular}{l|ccc|cc} 
 & \multicolumn{3}{c}{DensePose-Chimps} & \multicolumn{2}{c}{Chimp\&See} \\
\midrule
 $k$ &$\textbf{AP}_{DensePose}$  & $\textbf{AP}_{D}$ & $\textbf{AP}_{S}$ &  $\textbf{AP}_{D}$& $\textbf{AP}_{S}$
 \\
 \midrule
0 &  $33.8 \pm .2$ & $63.1 \pm .2$ & $57.9 \pm .2$ & $59.0 \pm .3$ & $49.2 \pm .4$ \\
$1$ &	$34.7 \pm .5$ &	$63.0 \pm .2$ & $57.9 \pm .3$ & $\mathbf{59.3  \pm .3}$ & $49.3 \pm  .6$\\
$2$ &	$34.6 \pm .6$ & $63.4 \pm .3$ & $57.9 \pm .3$ & $59.2 \pm  .4$ & $49.3 \pm .4$ \\ 
\textbf{5} &  $\mathbf{34.9 \pm .5}$ & $\mathbf{63.4 \pm .3}$ &	$\mathbf{58.0 \pm .2}$ & $59.2 \pm .4$ &	$49.2 \pm .5$ \\ %
$10$ &   $34.6 \pm .6$ & $63.3 \pm .3$ & $58.0 \pm .3$ & $59.2 \pm .4$ & $\mathbf{49.4 \pm .4}$\\ 
$1000$ & $33.1 \pm .6$ & $63.2 \pm .2$ & $57.8 \pm .3$ &  $59.2 \pm .5$ & $49.4 \pm .5$\\
$10000$ &  $27.6 \pm 4.6$ & $60.2 \pm .4$ & $55.7 \pm .5$ & $58.0 \pm .7$ & $49.1 \pm .6$\\
\bottomrule
\end{tabular}}
\caption{\label{selftraining_percent}DensePose, detection and instance segmentation AP of the \emph{student} network trained with $I$-sampling for different number of sampled points $k$. \textit{Mean}$\scriptstyle\pm$\textit{std} for $20$ runs.}
\end{table}

\subsection{Results}

\paragraph{Ablations on architectural choices.}
First, we compare our model to the original DensePose-RCNN~\cite{guler2018densepose} (detectron2 implementation). We also ablate our improvements in the architecture and provide results with and without auto-calibration.
Tab.~\ref{tab:auto_calibrated_dp_rcnn}, ~\ref{tab:auto_calibrated_dp_rcnn_2} show consistent improvements on all tasks for both modifications.


\begin{table}[!t]
\centering
\scalebox{0.72}{
\begin{tabular}{l|ccc} 
 selected COCO object classes & $\textbf{AP}$ & $\textbf{AP}_{50}$ & $\textbf{AP}_{75}$ 
 \\
 \midrule
 top-9 classes& 57.29 &		85.63 &	63.45\\
 bear-only & 40.69 &	70.88 &	44.23 \\
 person-only & 9.39 &		19.32 &	8.21\\

 animals-only & 52.28	&		80.62 &	58.60 \\
 person + animals & \textbf{57.34}	&		\textbf{85.76} &	\textbf{63.59}\\
 \midrule
 person + animals: class agnostic & 57.34	&	85.76 &	63.59\\
 person + animals: class specific & 50.47	&	72.85 & 	54.30 \\
\bottomrule
\end{tabular}}
\caption{\label{t:transfer} Instance segmentation AP on DensePose-Chimps for Mask R-CNN trained on different subsets of classes. 
}
\end{table}

\begin{figure*}[!t]
    \centering
    \vspace{-.75em}
    \includegraphics[width=0.822\linewidth]{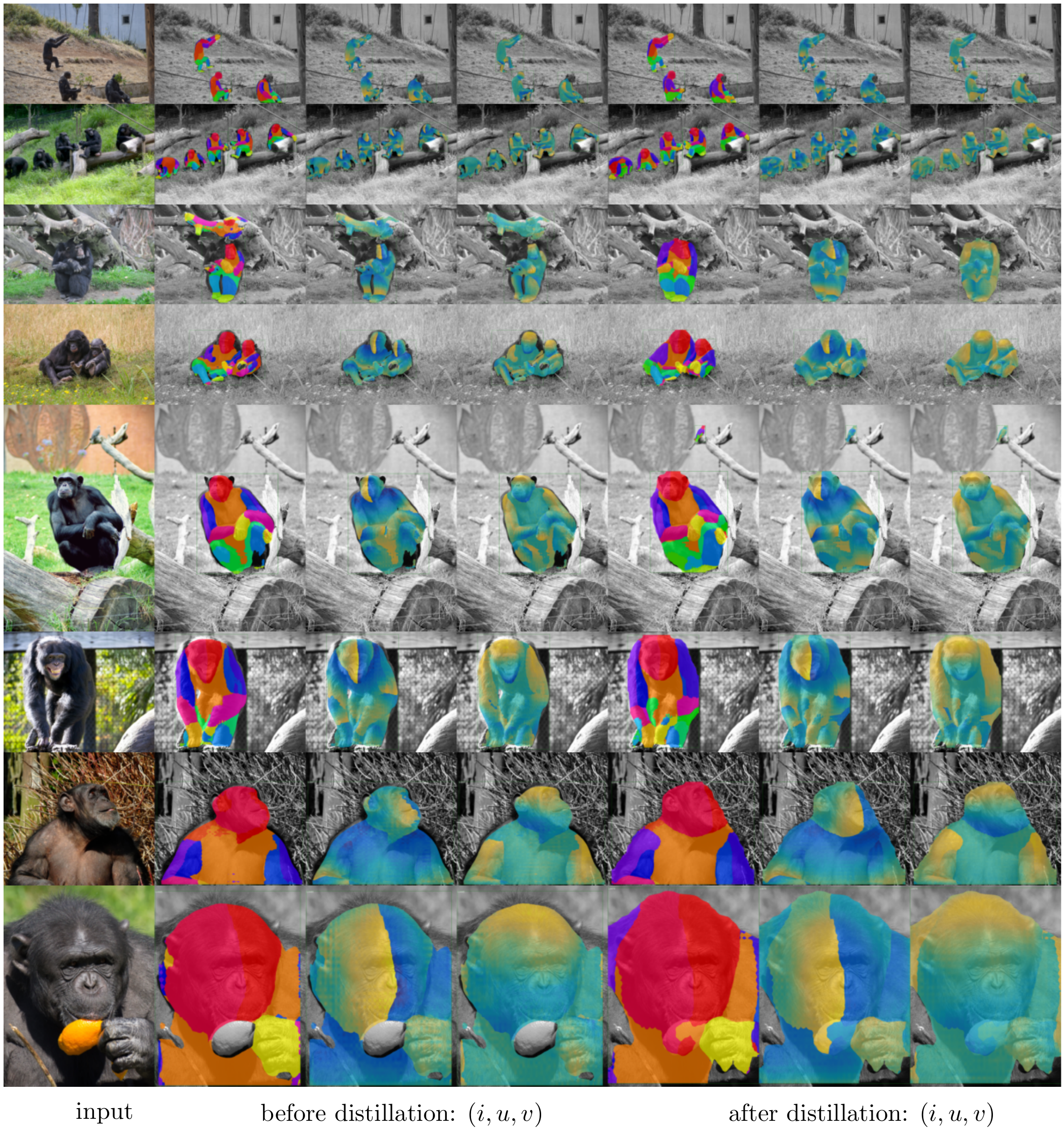}
    \vspace{-.5em}
    \caption{\label{fig:results} Visual results: \textbf{(left)} \emph{teacher} network predictions vs (\textbf{right}) predictions of \emph{student} network trained using $I$-sampling. The \emph{student} produces more accurate boundaries and $uv$-maps. Zoom-in for details.}
\end{figure*}


\paragraph{Optimal transfer support.}
\label{s:transferexp}
We (a) benchmarked every strategy for class selection described in Sect.~\ref{s:transfer} and (b) experimented with multi-class and class-agnostic models. 
From Tab.~\ref{t:transfer} we can see that class agnostic training on the \emph{animals+person} subset shows the best transferability for DensePose-Chimps dataset. Therefore, it was used for training all our DensePose models.

\paragraph{Dense label distillation.}
We conducted experiments with different sampling strategies and different numbers of sampled points $k$ per detection. In Tab.~\ref{selftraining_strategy} we show performance of the \emph{teacher} (first row) and the \emph{student} networks trained using different sampling strategies along with the corresponding optimal $k$.
$I$-based sampling showed most impressive gains, followed by $uv$-based sampling. Uniform selection produces poor results.
In Tab.~\ref{selftraining_percent} we report performance for different number of sampled points in every detection for $I$-based sampling. Qualitative results are shown in Fig.~\ref{fig:results}.

\section{Conclusions}\label{s:conclusions}

We have studied the problem of extending dense body pose recognition to animal species and suggested that doing this at scale requires learning from unlabelled data.
Encouragingly, we have demonstrated that existing detection, segmentation, and dense pose labelling models can transfer very well to a proximal animal class such as chimpanzee despite significant inter-class differences.
We have shown that substantial improvements can be obtained by carefully selecting which categories to use to pre-train the model, by using a class-agnostic architecture to integrate different sources of information, and by modelling labelling uncertainty to grade pseudo-label for self-training.
In this manner, we have been able to achieve excellent performance without using a single labelled image of the target class for training.

In the future, we would like to investigate how a limited amount of target supervision can be best used to improve the results, and how other techniques from domain adaptation could also be used for this purpose.

\newpage
\section{Acknowledgements}
We thank all parties performing or supporting collection of the Chimp\&See dataset, including:
\begin{itemize}
    \item [(a)] individual contributors: Theophile Desarmeaux, Kathryn J. Jeffery, Emily Neil, Emmanuel Ayuk Ayimisin, Vincent Lapeyre, Anthony Agbor, Gregory Brazzola, Floris Aubert, Sebastien Regnaut, Laura Kehoe, Lucy D’Auvergne, Nuria Maldonado, Anthony Agbor, Emmanuelle Normand, Virginie Vergnes, Juan Lapuente, Amelia Meier, Juan Lapuente, Alexander Tickle, Heather Cohen, Jodie Preece, Amelia Meier, Juan Lapuente, Roman M. Wittig, Dervla Dowd, Sorrel Jones, Sergio Marrocoli, Vera Leinert, Charlotte Coupland, Villard Ebot Egbe, Anthony Agbor, Volker Sommer, Emma Bailey, Andrew Dunn, Inaoyom Imong, Emmanuel Dilambaka, Mattia Bessone, Amelia Meier, Crickette Sanz, David Morgan, Aaron Rundus, Rebecca Chancellor, Felix Mulindahabi, Protais Niyigaba, Chloe Cipoletta, Michael Kaiser, Kyle Yurkiw, Bradley Larson, Alhaji Malikie Siaka, Liliana Pacheco, Manuel Llana, Henk Eshuis, Erin G. Wessling, Mohamed Kambi, Parag Kadam, Alex Piel, Fiona Stewart, Katherine Corogenes, Klaus Zuberbuehler, Kevin Lee, Samuel Angedakin, Kevin E. Langergraber, Christophe Boesch, Hjalmar Kuehl, Mimi Arandjelovic, Paula Dieguez, Mizuki Murai, Yasmin Moebius, Joana Pereira, Silke Atmaca, Kristin Havercamp, Nuria Maldonado, Colleen Stephens; 
    \item [(b)] funding agencies: Max Planck Society, Max Planck Society Innovation Fund, Heinz L. Krekeler Foundation; 
     \item [(c)] ministries and governmental organizations: Agence Nationale des Parcs Nationaux (Gabon), Centre National de la Recherche Scientifique (CENAREST) (Gabon), Conservation Society of Mbe Mountains (CAMM) (Nigeria), Department of Wildlife and Range Management (Ghana), Direction des Eaux, Forêts et Chasses (Senegal), Eaux et Forets (Mali), Forestry Commission (Ghana), Forestry Development Authority (Liberia), Institut Congolais pour la Conservation de la Nature (DR-Congo), Instituto da Biodiversidade e das Áreas Protegidas (IBAP), Makerere University Biological Field Station (MUBFS) (Uganda), Ministere de l’Economie Forestiere (R-Congo), Ministere de la Recherche Scientifique et de l'Innovation (Cameroon), Ministere de la Recherche Scientifique (DR-Congo), Ministere de l'Agriculture de l'Elevage et des Eaux et Forets (Guinea), Ministere de le Recherche Scientifique et Technologique (R-Congo), Ministere des Eaux et Forets (Cote d’Ivoire), Ministere des Forets et de la Faune (Cameroon), Ministre de l'Environnement et de l'Assainissement et du Developpement Durable du Mali, Ministro da Agricultura e Desenvolvimento Rural (Guinea-Bissau), Ministry of Agriculture, Forestry and Food Security (Sierra Leone), Ministry of Education (Rwanda), National Forestry Authority (Uganda), National Park Service (Nigeria), National Protected area Authority (Sierra Leone), Rwanda Development Board (Rwanda), Société Equatoriale d’Exploitation Forestière (SEEF) (Gabon), Tanzania Commission for Science and Technology (Tanzania), Tanzania Wildlife Research Institute (Tanzania), Uganda National Council for Science and Technology (UNCST), (Uganda), Uganda Wildlife Authority (Uganda); 
     \item [(d)] non-governmental organizations: Budongo Conservation Field Station (Uganda), Ebo Forest Research Station (Cameroon), Fongoli Savanna Chimpanzee Project (Senegal), Foundation Chimbo (Boe), Gashaka Primate Project (Nigeria), Gishwati Chimpanzee Project (Rwanda), Goualougo Triangle Ape Project, Jane Goodall Insitute Spain (Dindefelo) (Senegal), Korup Rainforest Conservation Society (Cameroon), Kwame Nkrumah University of Science and Technology (KNUST) (Ghana), Loango Ape Project (Gabon), Lukuru Wildlife Research Foundation (DRC), Ngogo Chimpanzee Project (Uganda), Nyungwe-Kibira Landscape, Rwanda-Burundi (WCS), Projet Grands Singes, La Belgique, Cameroon (KMDA), Station d’Etudes des Gorilles et Chimpanzees (Gabon), Tai Chimpanzee Project (Cote d’Ivoire), The Aspinall Foundation, (Gabon), Ugalla Primate Project (Tanzania), WCS (Conkouati-Douli NP) (R-Congo), WCS Albertine Rift Programme (DRC), Wild Chimpanzee Foundation (Cote d'Ivoire), Wild Chimpanzee Foundation (Guinea), Wild Chimpanzee Foundation (Liberia), Wildlife Conservation Society (WCS) Nigeria (Nigeria), WWF (Campo Ma’an NP) (Cameroon), WWF Congo Basin (DRC).
\end{itemize}

\let\oldthebibliography\thebibliography
\let\endoldthebibliography\endthebibliography
\renewenvironment{thebibliography}[1]{
  \begin{oldthebibliography}{#1}
    \setlength{\itemsep}{0em}
    \setlength{\parskip}{0em}
}
{
  \end{oldthebibliography}
}
{\small\bibliographystyle{ieee_fullname}\bibliography{refs}}

\clearpage

\appendix
\noindent{\LARGE \textbf{Appendix}}
\vspace{12pt}

In Section~\ref{sec:architecture} we provide more details on our implementation of the Multi-head R-CNN network. Then, in Section~\ref{sec:ablations} we describe additional ablation studies on the advantages of the auto-calibrated training, as well as other architectural choices. Finally, Section~\ref{sec:results} refers the reader to the qualitative results obtained on videos from the Chimp\&See dataset.

\section{Architecture}
\label{sec:architecture}
We introduced a number of changes and improvements in the DensePose head of the standard DensePose R-CNN architecture of~\cite{guler2018densepose} with ResNet-50 \cite{resnet} backbone.
These changes are listed below for the affected branches; other branches remained unchanged and correspond exactly to the Mask R-CNN architecture~of~\cite{he17mask}.
\begin{itemize}
    \item We have increased the RoI resolution from $14\times14$ to $28\times28$ in the DensePose head, as proposed in~\cite{yang2018parsing}.
    \item We have replaced the 8-layer DensePose head with the geometric and context encoding (GCE) module~\cite{yang2018parsing}, combining a non-local convolutional layer~\cite{wang2018nonlocal} with the atrous spatial pyramid pooling (ASPP)~\cite{chen2017encoder}.
    \item We have replaced the original FPN of DensePose R-CNN with a Panoptic FPN~\cite{kirillov2019panoptic}.
\end{itemize}
Each of these modifications led to increase in network performance due to improved multi-scale context aggregation. We refer the reader to the work of~\cite{yang2018parsing} for ablation studies whose results are aligned well with our own observations.

To predict α or σ we
simply extend the output layer of the corresponding head by
doubling the number of its neurons.

Our codebase, network configuration files for each experiment and pretrained models will be publicly released. 

\begin{table*}[t!]
\centering
\scalebox{1.0}{
\begin{tabular}{l|ll|ll|ll} 
 \multicolumn{1}{c}{\phantom{}}  & \multicolumn{2}{c}{COCO \texttt{minival}} & \multicolumn{2}{c}{DensePose-Chimps} & \multicolumn{2}{c}{Chimp\&See} \\
\midrule
model  & $\textbf{AP}_{D}$ & $\textbf{AP}_{S}$ &  $\textbf{AP}_{D}$& $\textbf{AP}_{S}$ & $\textbf{AP}_{D}$& $\textbf{AP}_{S}$
 \\
 \midrule



 Mask RCNN & 40.98 &	\textbf{37.17} &	48.3 &	44.92 &	40.56 &	33.91 \\
$\sigma$-Mask RCNN & \textbf{41.12} ( \pos{+0.14} ) &	37.09 ( \negative{-0.08} ) &	\textbf{52.05} ( \pos{+3.75} ) &	\textbf{47.94} ( \pos{+3.02} ) &	\textbf{42.9} ( \pos{+2.34} ) &	\textbf{34.74} ( \pos{+0.82} ) \\

\bottomrule
\end{tabular}}
\caption{\label{tab:maskrcnn}\textbf{Auto-calibrated Mask R-CNN \cite{he17mask}:} detection, instance segmentation on COCO \texttt{minival} (all classes).}
\end{table*}

\section{Computational cost}
Our auto-calibrated model has a negligible computational overhead ($< 1\%$) compared to the baseline model.
Before training the \emph{student}, sampling of the pseudo-labels requires one forward pass of the \emph{teacher} network
over the unlabeled dataset. The \emph{teacher} and the \emph{student} networks share the same architecture.

\section{Ablation studies}
\label{sec:ablations}

First, we report performance of the original Mask R-CNN~\cite{he17mask} framework, as well as our auto-calibrated version of the same architecture, on detection and segmentation tasks (see Tab.~\ref{tab:maskrcnn}). Training in the auto-calibration setting resulted in minor gains on the COCO dataset that the model was trained on, but, as expected, led to major improvements in performance on the out-of-distribution data (DensePose-Chimps and Chimp\&See). 

Second, Tab.~\ref{tab:masks} shows results of replacing the proposed binary foreground-background segmentation in the DensePose head (a) with 15-way coarse body part segmentation as in the original DensePose-RCNN framework~\cite{guler2018densepose} (b).
We can see that binary segmentation generalizes better than the 15-way.
We have also experimented with using the binary mask from the Mask R-CNN head instead of mask produced by the DensePose head (Tab.~\ref{tab:masks} (c)) \emph{during inference step}.
Moreover, even though exploiting the mask from the separate mask head at test time results in better performance, complete removal of the mask from the DensePose head leads to under-training and decreased accuracy of estimation of $uv$-coordinates (since in this case the DensePose head receives only sparse supervisory signals at the annotated locations).

\begin{table*}[!h]
\centering
\scalebox{1.0}{
\begin{tabular}{l|l l|lll} 
& model  & \scalebox{0.85}{Mask in DensePose head}  & $\textbf{AP}$ & $\textbf{AP}_{50}$ &  $\textbf{AP}_{75}$
 \\
 \midrule


a) & DensePose-RCNN* ($\sigma$) &  binary & 53.20 &				88.27 &	56.98 \\
b) & DensePose-RCNN* ($\sigma$) & 15-way & 50.87 &				86.91 &	54.49 \\
\midrule
c) & DensePose-RCNN* ($\sigma$) 
 + mask from the mask head 
& binary & \textbf{54.35} &	\textbf{88.58} & \textbf{60.28} \\
\bottomrule
\end{tabular}}
\caption{\label{tab:masks} \textbf{Ablation study of the mask in the DensePose head.} Reports the DensePose performance on DensePose-COCO \texttt{minival}.
a) our proposed architecture;
b) replace the binary segmentation of the DensePose head with 15-way coarse body part segmentation as in the original DensePose-RCNN framework~\cite{guler2018densepose};
c) use the binary mask from the DensePose head during training, but substitute it with the mask from the separate mask head during inference.}
\end{table*}

\section{Qualitative results}
\label{sec:results}
In addition, we also point the readers to the attached video samples from the Chimp\&See dataset showing frame-by-frame predictions produced by our model before (\emph{teacher}) and after self-training (\emph{student}).
The results produced by the \emph{student} network are generally significantly more stable.

\end{document}